\pdfoutput=1

\documentclass[11pt]{article}

\usepackage[]{ACL2023}
\usepackage{tikz}
\usepackage{fontawesome}
\usepackage{booktabs}
\usepackage{multirow}
\usepackage{amsfonts}
\usepackage{amsmath}
\usepackage{cleveref}
\usepackage{xcolor}
\usepackage{wasysym}
\usepackage{pgfkeys}

\usepackage{times}
\usepackage{latexsym}
\usepackage{pgfplots}
\usepackage{graphicx}
\usepackage{subcaption}
\usepackage[T1]{fontenc}

\usepackage[utf8]{inputenc}

\usepackage{microtype}

\usepackage{inconsolata}
\definecolor{freeze}{HTML}{4393c3}
\newcommand{\A}[0]{\mathbf{A}}
\newcommand{\B}[0]{\mathbf{B}}
\newcommand{\Atie}[0]{\A_{_{\text{\faChain}}}}
\newcommand{\Btie}[0]{\B_{_{\text{\faChain}}}}
\newcommand{\Atiefzn}[0]{\begingroup\color{freeze}\A_{_{\text{\faChain}}}\endgroup}
\newcommand{\Btiefzn}[0]{\begingroup\color{freeze}\B_{_{\text{\faChain}}}\endgroup}
\newcommand{\vvec}[0]{\mathbf{v}}
\newcommand{\uvec}[0]{\mathbf{u}}
\newcommand{\vvecfzn}[0]{\begingroup\color{freeze}\vvec\endgroup}
\newcommand{\uvecfzn}[0]{\begingroup\color{freeze}\uvec\endgroup}
\newcommand{\bluev}[0]{$\begingroup\color{freeze}\vvec\endgroup$}
\newcommand{\blueu}[0]{$\begingroup\color{freeze}\uvec\endgroup$}

\newcommand{\tiedlora}[0]{Tied-LoRA}
\newcommand{\Tiedlora}[0]{Tied-LoRA}
\newcommand{\TiedLora}[0]{Tied-LoRA}

\newcommand{\lorasym}[0]{$\vvecfzn\B\uvecfzn\A$}
\newcommand{\lora}[0]{LoRA~($\vvecfzn\B\uvecfzn\A$)~}
\newcommand{\Lora}[0]{LoRA~($\vvecfzn\B\uvecfzn\A$)~}

\newcommand{\verasym}[0]{$\vvec\Btiefzn\uvec\Atiefzn$}
\newcommand{\Tuv}[0]{Vera~($\vvec\Btiefzn\uvec\Atiefzn$)~}
\newcommand{\TA}[0]{TL\textsubscript{1}($\vvecfzn\Btiefzn\uvecfzn\Atie$)~}
\newcommand{\TAuv}[0]{TL\textsubscript{2}($\vvec\Btiefzn\uvec\Atie$)~}
\newcommand{\TB}[0]{TL\textsubscript{3}($\vvecfzn\Btie\uvecfzn\Atiefzn$)~}
\newcommand{\TBu}[0]{TL\textsubscript{4}($\vvecfzn\Btie\uvec\Atiefzn$)~}
\newcommand{\TAB}[0]{TL\textsubscript{5}($\vvecfzn\Btie\uvecfzn\Atie$)~}
\newcommand{\TABuv}[0]{TL\textsubscript{6}($\vvec\Btie\uvec\Atie$)~}

\newcommand{\LineTAB}[0]{TL\textsubscript{5}\\($\vvecfzn\Btie\uvecfzn\Atie$)~}

\pgfplotsset{compat=1.3}

\title{\tiedlora: Enhancing parameter efficiency of LoRA with Weight Tying}

\author{Adithya Renduchintala \\\And
  Tugrul Konuk \\
  \\
  NVIDIA \\
  \texttt{\{adithyare,tkonuk,okuchaiev\}@nvidia.com} \\\And 
  Oleksii Kuchaiev\\
  }

\begin{document}
\maketitle
\begin{abstract}
We introduce \tiedlora, a novel paradigm leveraging weight tying and selective training to enhance the parameter efficiency of Low-rank Adaptation (LoRA). Our exploration encompasses different plausible combinations of parameter training and freezing, coupled with weight tying, aimed at identifying the optimal trade-off between performance and the count of trainable parameters. Across 
$5$ diverse tasks and two foundational language models with different parameter counts, our experiments provide comprehensive insights into the inherent trade-offs between efficiency and performance.

Our findings reveal a specific \tiedlora~configuration that distinguishes itself by showcasing comparable performance to LoRA across multiple tasks while utilizing only a fraction of the parameters employed by the standard LoRA method, particularly at elevated ranks. This underscores the efficacy of \tiedlora~in achieving impressive results with significantly reduced model complexity. 
\end{abstract}

\section{Introduction}
Large language models (LLMs) play a crucial role in various Natural Language Processing (NLP) applications due to their proficiency. A significant factor driving their widespread adoption is the ability to fine-tune pretrained LLMs efficiently for specific downstream tasks. This fine-tuning process allows the creation of specialized language models that excel in specific domains and tasks. Despite dealing with smaller training data compared to pretraining, the computational demand for during fine-tuning remains high, especially for large models with billions of parameters.

Moreover, for LLM service providers, it is often necessary to cater to diverse requirements by maintaining distinct customizations for each combination of user and task in the service. For instance, consider a scenario where a language model is employed to assist users in generating content for social media. User X may have preferences for formal language and professional tone, while another user, Y, might prefer a more casual and conversational style. Additionally, each user may have different tasks, such as composing business emails, translating documents, creating social media captions or drafting blog posts. To serve these varied preferences and tasks simultaneously, the language model needs to be finely tuned for each specific combination of user (X or Y) and task (email composition or translation).

  
\tikzset {_qp3twg8z4/.code = {\pgfsetadditionalshadetransform{ \pgftransformshift{\pgfpoint{0 bp } { 0 bp }  }  \pgftransformrotate{0 }  \pgftransformscale{2 }  }}}
\pgfdeclarehorizontalshading{_y56jsd4cl}{150bp}{rgb(0bp)=(0.96,0.65,0.14);
rgb(37.5bp)=(0.96,0.65,0.14);
rgb(62.5bp)=(0.29,0.56,0.89);
rgb(100bp)=(0.29,0.56,0.89)}

  
\tikzset {_ulzxcji3v/.code = {\pgfsetadditionalshadetransform{ \pgftransformshift{\pgfpoint{0 bp } { 0 bp }  }  \pgftransformrotate{0 }  \pgftransformscale{2 }  }}}
\pgfdeclarehorizontalshading{_9u7wc044m}{150bp}{rgb(0bp)=(0.96,0.65,0.14);
rgb(37.5bp)=(0.96,0.65,0.14);
rgb(62.5bp)=(0.29,0.56,0.89);
rgb(100bp)=(0.29,0.56,0.89)}

  
\tikzset {_fyo5tisnq/.code = {\pgfsetadditionalshadetransform{ \pgftransformshift{\pgfpoint{0 bp } { 0 bp }  }  \pgftransformrotate{0 }  \pgftransformscale{2 }  }}}
\pgfdeclarehorizontalshading{_xbeccsso4}{150bp}{rgb(0bp)=(0.96,0.65,0.14);
rgb(37.5bp)=(0.96,0.65,0.14);
rgb(62.5bp)=(0.29,0.56,0.89);
rgb(100bp)=(0.29,0.56,0.89)}

  
\tikzset {_h6rlcfism/.code = {\pgfsetadditionalshadetransform{ \pgftransformshift{\pgfpoint{0 bp } { 0 bp }  }  \pgftransformrotate{0 }  \pgftransformscale{2 }  }}}
\pgfdeclarehorizontalshading{_ye7qha5sz}{150bp}{rgb(0bp)=(0.96,0.65,0.14);
rgb(37.5bp)=(0.96,0.65,0.14);
rgb(62.5bp)=(0.29,0.56,0.89);
rgb(62.5bp)=(0.29,0.56,0.89);
rgb(100bp)=(0.29,0.56,0.89)}
\tikzset{every picture/.style={line width=0.75pt}} 
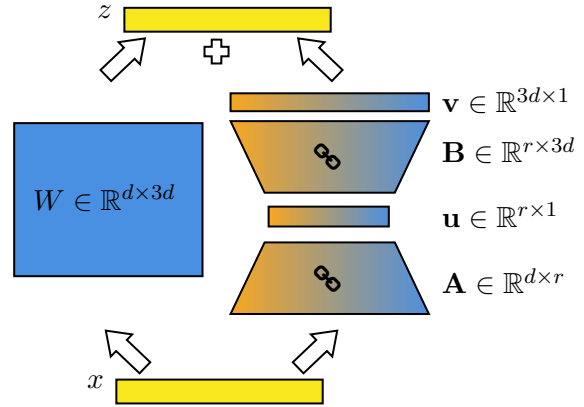
\begin{figure}	
\centering
\begin{tikzpicture}[x=0.75pt,y=0.75pt,yscale=-1,xscale=1]

\path  [shading=_y56jsd4cl,_qp3twg8z4] (225,76) -- (208,112) -- (143,112) -- (126,76) -- cycle ; 
 \draw   (225,76) -- (208,112) -- (143,112) -- (126,76) -- cycle ; 

\path  [shading=_9u7wc044m,_ulzxcji3v] (145,119) -- (205,119) -- (205,129) -- (145,129) -- cycle ; 
 \draw   (145,119) -- (205,119) -- (205,129) -- (145,129) -- cycle ; 

\draw  [fill={rgb, 255:red, 74; green, 144; blue, 226 }  ,fill opacity=1 ] (18,77) -- (112,77) -- (112,154) -- (18,154) -- cycle ;
\draw  [fill={rgb, 255:red, 248; green, 231; blue, 28 }  ,fill opacity=1 ] (73,19) -- (176,19) -- (176,31) -- (73,31) -- cycle ;
\draw  [fill={rgb, 255:red, 248; green, 231; blue, 28 }  ,fill opacity=1 ] (69,206) -- (172,206) -- (172,218) -- (69,218) -- cycle ;
\draw   (116.6,35) -- (121.4,35) -- (121.4,38.6) -- (125,38.6) -- (125,43.4) -- (121.4,43.4) -- (121.4,47) -- (116.6,47) -- (116.6,43.4) -- (113,43.4) -- (113,38.6) -- (116.6,38.6) -- cycle ;
\draw   (157.08,197.27) -- (168.78,186.06) -- (166.11,183.28) -- (179.25,181.37) -- (176.79,194.42) -- (174.12,191.63) -- (162.42,202.84) -- cycle ;
\draw   (85.92,197.27) -- (74.22,186.06) -- (76.89,183.28) -- (63.75,181.37) -- (66.21,194.42) -- (68.88,191.63) -- (80.58,202.84) -- cycle ;
\draw   (181.92,50.27) -- (170.22,39.06) -- (172.89,36.28) -- (159.75,34.37) -- (162.21,47.42) -- (164.88,44.63) -- (176.58,55.84) -- cycle ;
\draw   (61.08,50.27) -- (72.78,39.06) -- (70.11,36.28) -- (83.25,34.37) -- (80.79,47.42) -- (78.12,44.63) -- (66.42,55.84) -- cycle ;
\path  [shading=_xbeccsso4,_fyo5tisnq] (225,173) -- (208,137) -- (143,137) -- (126,173) -- cycle ; 
 \draw   (225,173) -- (208,137) -- (143,137) -- (126,173) -- cycle ; 

\path  [shading=_ye7qha5sz,_h6rlcfism] (126,62) -- (225,62) -- (225,71) -- (126,71) -- cycle ; 
 \draw   (126,62) -- (225,62) -- (225,71) -- (126,71) -- cycle ; 

\draw (26,106.36) node [anchor=north west][inner sep=0.75pt]  {$W\in\mathbb{R}^{d\times 3d}$};
\draw (230,148.4) node [anchor=north west][inner sep=0.75pt] {$\A\in\mathbb{R}^{d\times r}$};
\draw (166,147) node [anchor=north west][inner sep=0.75pt] {\faChain};
\draw (230,57.4) node [anchor=north west][inner sep=0.75pt] {$\vvec\in\mathbb{R}^{3d\times 1}$};
\draw (230,82.4) node [anchor=north west][inner sep=0.75pt]  {$\B\in\mathbb{R}^{r\times 3d}$};
\draw (166,86) node [anchor=north west][inner sep=0.75pt]  {\faChain};
\draw (53,202.4) node [anchor=north west][inner sep=0.75pt] {$x$};
\draw (58,16.4) node [anchor=north west][inner sep=0.75pt] {$z$};
\draw (230,115.4) node [anchor=north west][inner sep=0.75pt] {$\uvec\in\mathbb{R}^{r\times 1}$};

\end{tikzpicture}
	\caption{Schematic of our Tied-Lora paradigm, the main low-rank matrices $A$ and $B$ are tied across (indicated by the {\small\faChain} symbol) all the layers of the base language model. 
 We use the gradient shading to indicate that these parameters can either be trained or frozen.
	}\label{fig:setup}
\end{figure}

As the number of users and tasks per user increases, so does the complexity and cost associated with customization. Managing and storing the various combinations of customizations for each user-task pair can introduce additional expenses, especially after the initial training phase. The storage and retrieval of these customized models, each tailored to specific user preferences and tasks, contribute to ongoing operational costs. Therefore, in addition to efficient utilization of customizable parameters during training, careful consideration must be given to the post-training phase, where the cost of saving and accessing these combinations becomes a significant factor in the overall resource management strategy. This holistic approach is crucial for maintaining optimal performance across diverse user-task combinations while keeping both computational and operational costs in check.

In light of these challenges, developing effective customization methods that not only enhance model performance but also reduce the number of training parameters becomes crucial. Parameter-efficient fine-tuning (PEFT) emerges as a valuable approach in this context. PEFT involves refining pretrained models with minimal parameter updates, enabling the creation of specialized models that excel in specific domains and tasks. This streamlined customization process not only optimizes parameter utilization during training but also mitigates the costs associated with managing, storing, and serving diverse customizations post-training. 


Low-rank Adaptation (LoRA) method~\citep{hu2021lora}, stands out as a popular and efficient parameter-efficient fine-tuning (PEFT) approach, offering a straightforward implementation and the ability to integrate LoRA weights into the base model post-training. 
Despite its advantages, the expense of LoRA training becomes more pronounced, particularly with the growing size of base language models. 
While previous efforts focused on enhancing LoRA's parameter efficiency through careful low-rank selection, we introduce an alternative approach. 
In contrast to controlling parameter count through rank, our method incorporates simple weight tying alongside selective training. 
This novel combination forms the basis for a range of \tiedlora~configurations, each evaluated for performance across five diverse customization tasks. 
Through this approach, we aim to push the boundaries of parameter-efficient fine-tuning, making advancements in both effectiveness and simplicity.
Our contributions are threefold:
\begin{enumerate}
    \item We propose a range of \tiedlora~configurations that use simple weight tying in LoRA along with selective training to boost the parameter efficiency of LoRA.
    \item We study this spectrum of possible \tiedlora~configurations on diverse tasks that resemble real-world customization problems.
    \item Based on the results of our study, we propose the specific \TABuv configuration as the best option for maintaining performance while reducing parameters. This configuration is within $1-2\%$ of LoRA in terms of performance and in one case beats LoRA while only using $12.5\%$ of the number of parameters.
\end{enumerate}
\section{Method}
In this section, we introduce \Tiedlora, a paradigm for parameter-efficient fine-tuning of large language models through low-rank weight-update approximations, weight-tying and selective training. 
Our framework offers a range of ``LoRA-like'' configurations through a series of design choices over selective parameter training and weight tying, including some of the existing PEFT methodologies available in the literature. 
Specifically, we use weight tying alongside pairs of projection matrices and scaling vectors that can be selectively either trained or frozen. 
As the low-rank computation path does not introduce any non-linearity, all \tiedlora~configurations can be merged into the base model weights to preventing additional latency during inference. 

\Cref{tab:methods} provides an overview of the scenarios we study.
We refer to each configuration in our study with TL (Tied-LoRA) followed by a subscript index (e.g., TL\textsubscript{1}).
Additionally, we also include the template of possible training parameters ($\mathbf{v},\mathbf{B},\mathbf{u}$ and $\mathbf{A}$, discussed in \cref{sec:formualation}).
For \Tiedlora, the low-rank projection matrices $\mathbf{A}$ and $\mathbf{B}$ are \emph{tied} across all the layers of the base model which we indicate using the subscript $_{_{\text{\faChain}}}$.
We also indicate if a parameter is frozen by ~\textcolor{freeze}{blue} font and a trainable parameter with regular font.
Thus, traditional LoRA can be expressed as \lorasym~and VeRA \citep{kopiczko2023vera} can be expressed as \verasym.
\subsection{Formulation}\label{sec:formualation}
The overall structure of the tied LoRA framework can be seen in \Cref{fig:setup}. 
Note that the original LoRA~\citep{hu2021lora} uses a dedicated pair of low-rank projections for each of the $\mathbf{Q}, \mathbf{K}, \mathbf{V}$ matrices. 
However, in our formulation, $W$ is a $d \times 3d$ matrix that jointly projects $\mathbf{Q}, \mathbf{K}$, and $\mathbf{V}$ attention matrices, where $d$ is the hidden size of the base language model. Therefore, our down projection $\mathbf{A}$ is a $d \times r$ shaped matrix and up projection matrix $\mathbf{B}$ has shape $r \times 3d$, where $r$ is the low-rank bottleneck dimension. Essentially, the down projection $\mathbf{A}$ is \emph{shared} by $\mathbf{Q}, \mathbf{K}$, and $\mathbf{V}$, leading to fewer trainable parameters ($4dr$) than the original LoRA ($6dr$).

For a linear layer with a frozen pretrained weight matrix $\mathbf{W}$, we define the layer output as
    \begin{equation}
        z = \mathbf{W}x + \Delta \mathbf{W} x \approx \mathbf{W}x + \frac{\alpha}{r} \Lambda_{v}\mathbf{B}\Lambda_{u}\mathbf{A}x,
        \label{eqn:gen_lora}
    \end{equation}
    where $\Delta \mathbf{W}$ is the full-rank update matrix, $\alpha$ is a scaling factor, $\mathbf{A}$ and $\mathbf{B}$ are low-rank projection matrices, and $\Lambda_{u}$ and $\Lambda_{v}$ are diagonal matrices with diagonal elements given by $u$ and $v$, respectively. Herein, $\Lambda_{v}\mathbf{B}\Lambda_{u}\mathbf{A}x$ is the low-rank approximation to the parameter update matrix $\Delta \mathbf{W}$. Unlike the original LoRA, where $\alpha$ is a hyper-parameter that can be manually set, we simply set $\alpha = r$, effectively removing its scaling effect. 

    Equation~\ref{eqn:gen_lora} is a generalized formulation for methods that utilize low-rank approximations to estimate parameter updates. Particular settings of parameter updates and weight tying reduces this equation to some of the existing formulations in the literature. Setting and freezing $\Lambda_{u}=\Lambda_{v}=I$ and untying $\mathbf{A}$ and $\mathbf{B}$ results in LoRA:
    \begin{equation}
        z = \mathbf{W} x + \mathbf{BA}x.
        \label{eqn:lora}
    \end{equation} 
    
    Similarly, randomly initializing $\mathbf{A}$ and $\mathbf{B}$ matrices and tying them across all layer leads the the VeRA formulation~\citep{kopiczko2023vera}:
\begin{equation}
    z = \mathbf{W}x + \Lambda_{v}\mathbf{B}\Lambda_{u}\mathbf{A}x,
    \label{eqn:vera}
    \end{equation}    

\begin{table}[]
\centering
\small{\resizebox{\columnwidth}{!}{%
\begin{tabular}{@{}llll@{}}
\toprule
Method & Parameters &  Initialization  \\ \midrule
\lora         & $4Ldr$          & $A \sim\mathcal{N}, B = 0, u,v = 1$  \\
\Tuv          & $L(r+3d)$       & $A,B \sim\mathcal{N}, u=1, v=0$ \\
\TA           & $dr$            & $A,B \sim\mathcal{N}, u,v=1$\\
\TAuv         & $dr+L(r+3d)$    & $A,B \sim\mathcal{N}, u=1, v=0$ \\
\TB           & $3dr$           & $A \sim\mathcal{N}, B = 0, u,v=1$\\
\TBu          & $(L+3d)r$       & $A \sim\mathcal{N}, B = 0, v,u=1$      \\
\TAB          & $4dr$           & $A \sim\mathcal{N}, B = 0, u,v = 1$  \\
\TABuv        & $4dr+L(r+3d)$   & $A,B \sim\mathcal{N}, u=1, v=0$  \\ \bottomrule
\end{tabular}
}}
\caption{\tiedlora~configurations included in our study.
The first column shows acronyms used to identify each \tiedlora~configuration (i.e., method).
Symbols with subscript $_{_{\text{\faChain}}}$ indicate that it is shared across all layers and the color~\textcolor{freeze}{blue} indicates that the parameter is frozen.
Formulas for the number of trainable parameters in each configuration as a function of number of layers $L$, hidden size $d$, and low-rank $r$ are also provided.}
\label{tab:methods}
\end{table}

\subsection{Weight Tying}
The third column of~\Cref{tab:methods} presents representations for number of trainable parameters each Tied-Lora configuration requires. As is apparent from the table, weight tying is a critical ingredient of our proposed approach which drastically reduces the number of trainable parameters. For example, \Lora training using the 7B LLaMA-2~\citep{touvron2023llama2} language model with a typical low rank setting of $8$ requires $\sim4.2$M trainable parameters. By merely introducing weight tying across the $32$ layers of this model reduces the trainable parameters to $\sim131$K, which is a $96.875\%$ reduction. In comparison, the Vera method results in a reduction of $90.6\%$.
\subsection{Selective Training}
Through the flexible framework that equation~\ref{eqn:gen_lora} offers, we are given the opportunity to investigate a range training configurations. By selectively updating the components $A, B, u$, and $v$ during the training process, we can generate a variety of methodological variations. These variations not only exhibit differences in parameter count, but they also demonstrate distinct capabilities across a variety of tasks. This exploration allows us to investigate the intriguing regime of extremely low-parameter and low-rank PEFT models. This is a key step towards the customization of models, enabling them to excel at specific tasks while maintaining a minimal parameter count. Our ultimate goal here is to harness the power of this methodology to create highly efficient, task-specific models that achieve high performance with reduced complexity.    
\section{Experiments}
We now turn to evaluating the different configurations possible within our \tiedlora~paradigm.
While \Lora and PEFT methods can be used to train models for general instruction following~\citep{sun2023comparative,lermen2023lora,sun2023comparative}, we focus our evaluations in a ``task customization'' perspective, where each model is trained on a specific task and is evaluated on a test set from the same task.
\subsection{Tasks \& Datasets}
To evaluate the performance of each \tiedlora~configuration across diverse data settings, we utilized the following types of tasks:
\paragraph{Extractive QA} is a common task where the model is expected to ``read'' some relevant text (the context) and answer questions. The answers are usually exact sub-strings from the provided context.
We use SQuADv1 dataset \citep{rajpurkar-etal-2016-squad} in our experiments. Since the official test split of this dataset does not contain ground-truth answers, we use the validation set as our test set. We create a validation set comprising of a random sample of $4800$ examples extracted from the training set.
\paragraph{Summarization} is a central problem in NLP and several variations of summarization datasets have been proposed.
We employ the DialogSum dataset~\citep{chen-etal-2021-dialogsum} to study our models' performance on this task.
DialogSum includes summaries of real-word conversations on a diverse set of topics and scenarios.
This dataset was an attractive option as the length of the conversations and summarizes are within the context lengths ($4096$ tokens) of the base language models.
\paragraph{Commonsense Natural Language Inference (NLI)} is a task designed to probe the ability of language models to apply ``commonsense reasoning'' to choose a possible ending for a given situation described in natural language. These tasks are typically trivial for humans but language models can still struggle. 
We use the HellaSwag dataset~\citep{zellers-etal-2019-hellaswag} to study the performance of our proposed models on this type of task.
As HellaSwag contains multiple-choice questions, it can be viewed as a classification problem.
\paragraph{Translation} Machine translation is a natural language generation task which is widely used in research and industry. 
Translation is inherently multilingual and thus offers a challenging domain to study our \tiedlora~paradigm. 
There are several large scale translation datasets but we focus on a moderately sized IWSLT 2017 German-to-English spoken language translation dataset~\citep{cettolo-etal-2017-overview}.
With over $206k$ training examples this is the largest dataset we study.
\paragraph{Mathematical Reasoning} is a challenging domain where large language models still lag behind human performance. 
Using PEFT methods on such tasks further amplifies these challenges as there are very few trainable parameters.
In our experiments, we use the GSM8K benchmark~\citep{cobbe2021training} which contains $8.5$K high-quality, grade-school level math word problems.
Each example in the GSM8K benchmark contains a question and an answer. 
The answers are provided with natural language solutions which contain explanations of each step used to obtain the final answer. The final numerical answer is demarcated from the rest of the natural language solution. 
We evaluate our models by comparing these final numerical answers. 
\begin{table*}[!htb]
\centering
\small{
\resizebox{\textwidth}{!}{
\begin{tabular}{@{}llrrrrrrrrrrrrrrr@{}}
\toprule
  \multicolumn{1}{c}{\begin{tabular}[c]{@{}c@{}}Base Model\end{tabular}} &
  \multicolumn{1}{c}{\begin{tabular}[c]{@{}c@{}}Method\end{tabular}} &
  \multicolumn{3}{c}{\begin{tabular}[c]{@{}c@{}}Dialogsum\end{tabular}} &
  \multicolumn{3}{c}{\begin{tabular}[c]{@{}c@{}}GSM8K\end{tabular}} &
  \multicolumn{3}{c}{\begin{tabular}[c]{@{}c@{}}HellaSwag\end{tabular}} &
  \multicolumn{3}{c}{\begin{tabular}[c]{@{}c@{}}IWSLT 2017\end{tabular}} &
  \multicolumn{3}{c}{\begin{tabular}[c]{@{}c@{}}Squad\end{tabular}} \\ 
   &  & RougeL & $r$ & P\% & EM & $r$ & P\% & Acc. & $r$ & P\% & BLEU & $r$ & P\% & EM & $r$ & P\% \\ \midrule
\multirow{7}{*}{LLaMA2 7B}& \lora       & 40.76&8&100  & 32.75&64&100 & 91.97&16&100 & 41.30&8&100   & 88.52&2&100 \\ \cmidrule(l){2-17} 
                            & \Tuv        & 38.77&8&9.4  & 27.22&64&1.2 & 89.91&16&4.7 & 40.22&8&9.4   & 87.69&2&37.5 \\
                            & \TA         & 38.73&8&0.8  & 27.07&64&0.8 & 90.03&16&0.8 & 40.34&8&0.8   & 87.72&2&0.8 \\
                            & \TAuv       & 38.69&8&10.2  & 27.07&64&2.0 & 90.11&16&5.5 & 40.35&8&10.2   & 87.67&2&38.3 \\
                            & \TB         & 40.20&8&2.3  & 17.74&64&2.3 & 89.38&16&2.3 & 39.93&8&2.3   & 87.34&2&2.3 \\
                            & \TBu        & 39.46&8&2.3  & 21.00&64&2.3 & 89.46&16&2.3 & 40.34&8&2.3   & 87.06&2&2.3 \\
                            & \TAB        & \textbf{40.62}&8&3.1  & 30.33&64&3.1 & \textbf{91.75}&16&3.1 & 40.01&8&3.1   & 87.11&2&3.1 \\
                            & \TABuv      & 39.24&8&12.5  & \textbf{31.77}&64&4.3 & 91.15&16&7.8 & \textbf{41.33}&8&12.5   & \textbf{87.97}&2&40.6 \\ \cmidrule(l){2-17} 
                            & \Tuv   & 40.07&64&9.4 & 29.11&16&1.2 & 90.47&2&4.7  & 40.41&16&9.4  & 87.69&2&37.5 \\
                            & \TA    & 39.74&16&1.6 & 29.95&16&0.2 & 90.52&4&0.2  & 40.52&64&6.3  & 87.72&2&0.8 \\
                            & \TAuv  & 39.81&64&15.7 & 28.73&16&1.4 & 90.32&2&4.8  & 40.50&128&22.0 & 87.67&2&38.2 \\
                            & \TB    & 40.20&8&2.3  & 24.34&4&0.1  & 90.27&8&1.2  & 40.48&16&4.7  & 87.62&8&9.4 \\
                            & \TBu   & 40.17&16&4.7 & 25.70&8&0.3  & 90.18&4&0.6  & 40.65&16&4.7  & 87.72&4&4.7 \\
                            & \TAB   & \textbf{40.62}&8&3.1  & 30.33&64&3.1 & 91.75&16&3.1 & \textbf{41.37}&16&6.3  & 88.22&4&6.3 \\
                            & \TABuv & 39.71&16&15.6 & \textbf{31.77}&64&4.3 & \textbf{91.90}&64&17.2 & \textbf{41.37}&32&21.9  & \textbf{88.49}&4&43.8 \\ \midrule
\multirow{7}{*}{GPT-2B-001} & \lora       & 38.59&4&100  & 12.28&64&100 & 85.64&64&100 & 40.19&128&100 & 83.58&32&100 \\ \cmidrule(l){2-17} 
                            & \Tuv        & 37.02&4&18.8  & 6.97&64&1.2  & 75.94&64&1.2 & 38.20&128&0.6 & 79.43&32&2.3 \\
                            & \TA         & 37.11&4&1.0  & 8.26&64&1.0  & 76.32&64&1.0 & 38.12&128&1.0 & 79.26&32&1.0 \\
                            & \TAuv       & 37.00&4&19.8  & 8.11&64&2.2  & 77.02&64&2.2 & 38.17&128&1.6 & 79.50&32&3.4 \\
                            & \TB         & 36.50&4&3.1  & 5.69&64&3.1  & 25.05&64&3.1 & 36.46&128&3.1 & 76.96&32&3.1 \\
                            & \TBu        & 36.82&4&3.1  & 6.82&64&3.1  & 25.05&64&3.1 & 32.98&128&3.1 & 77.47&32&3.1 \\
                            & \TAB        & 37.17&4&4.2  & 8.34&64&4.2  & 82.25&64&4.2 & 38.58&128&4.2 & 81.43&32&4.2 \\
                            & \TABuv      & \textbf{37.63}&4&22.9  & \textbf{9.78}&64&5.3  & \textbf{85.02}&64&5.3 & \textbf{39.74}&128&4.8 & \textbf{83.02}&32&6.5 \\ \cmidrule(l){2-17} 
                            & \Tuv   & 37.28&8&18.8  & 8.26&2&1.2   & 83.41&2&1.2  & 39.15&2&0.6   & 81.77&2&2.3  \\
                            & \TA    & 37.22&8&2.1  & 9.55&4&0.1   & 83.54&4&0.1  & 39.09&2&0.1   & 82.20&2&0.1  \\
                            & \TAuv  & 37.29&8&20.1  & 9.40&4&1.2   & 83.64&2&1.2  & 39.11&2&0.6   & 82.41&4&2.5  \\
                            & \TB    & 37.18&16&12.5 & 6.97&8&0.4   & 80.66&4&0.2  & 38.25&4&0.1   & 80.96&4&0.4  \\
                            & \TBu   & 36.88&32&25.1 & 7.20&32&1.6  & 80.51&4&0.2  & 38.30&8&0.2   & 81.03&8&0.8  \\
                            & \TAB   & 37.55&8&8.3  & 9.40&128&8.3 & 83.71&32&2.1 & 39.20&64&2.1  & 82.74&16&2.1 \\
                            & \TABuv & \textbf{37.81}&32&52.2 & \textbf{10.31}&16&2.2 & \textbf{85.13}&32&3.3 & \textbf{39.74}&128&4.8 & \textbf{83.56}
&64&10.8 \\ \bottomrule
\end{tabular}
}
}
\caption{The results entire spectrum of \TiedLora\ configurations on five tasks using LLaMA2 7B base model and the GPT-2B-001 base model. 
For each base model section, the first row shows the best \lora scores on each task along with rank $r$ at which the best score was achieved. 
}
\label{tab:result}
\end{table*}
\subsection{Base Language Models}
Although PEFT enables the base language model to perform new tasks, the final performance heavily depends on the inherent abilities learned during pretraining.
This necessitates investigating the performance of \tiedlora~on multiple base models with different inherent capabilities.
Therefore, we use a relatively small two billion parameter, GPT-2B-001 decoder-only model\footnote{\url{https://huggingface.co/nvidia/GPT-2B-001}} released by NVIDIA and the moderately large $7$B LLaMA 2 model~\citep{touvron2023llama2} released by Meta. 
Additionally, these models also differ in the amount of pretraining data used. 
The GPT-2B-001 model was trained on $1.1$ trillion tokens of text from publicly available multilingual text spanning $53$ languages.
The LLaMA2 $7$B model was trained on $2$ trillion tokens of predominately English text. Both models are auto-regressive language models with a context size of $4096$ tokens.
\subsection{Implementation Details}
We use the open-source NeMo Framework to implement all the algorithms presented in this paper. 
Our implementation is publicly available through the NeMo GitHub repository.\footnote{\url{https://github.com/NVIDIA/NeMo/tree/adithyare/vera}} 
We set max training steps to $2k$, but training was terminated sooner using early stopping with a patience of $10$ to prevent over-fitting.
We trained all configurations using AdamW optimizer~\citep{loshchilov2017decoupled} with a weight decay of $0.01$ and a cosine learning rate schedule with $50$ warm-up steps.

For each Tied-Lora method we tried two learning rates, a high rate of $1^{-4}$ and a low learning rate of $1^{-5}$.
While the ``typical'' range of the low-rank dimension $r$ is $4-16$ we find that some complex tasks benefit from higher $r$ so we trained all our models with a wide range of $r \in \{2,4,8,\ldots,128\}$. 
Each task was trained with a global batch size of $256$ and a validation check interval of $30$ steps. 
The only exception was the IWSLT translation dataset for which we set global batch size and validation check interval of $1024$ and $60$ respectively.
No extensive hyper-parameter search was conducted.
We used greedy-decoding to generate the models' predictions with a limit of $500$ tokens.
\section{Results}
\Cref{tab:result} provides a detailed comparison  of various \TiedLora~configurations across our $5$ tasks for the LLaMA2 7B and GPT-2B-001 base models.
For each task we report the metric used, such as RougeL \citep{lin-och-2004-automatic}, Exact Match (EM), Accuracy and BLEU~\citep{papineni-etal-2002-bleu}, and the rank $r$ used.
For each model, the table is segmented into two sections: The first section compares the performance of all \TiedLora~configurations at the \emph{same rank} where \Lora achieved its optimum score. 
The second section shows the best performance of achieved by each configuration. 
In addition to  metric scores and rank ($r$) we also report parameter usage percentage (P\%) as a comparison to the parameter count of the best-performing Lora configuration. 
This offers a direct measure of efficiency, showing how each model, especially \TAB and \TABuv, leverages a smaller percentage of parameters compared to the \Lora for achieving its results. 

\begin{table*}[h!]
\centering
\small{%
\begin{tabular}{@{}lrrrrrrrrrc@{}}
\toprule
 & \multicolumn{9}{c}{LoRA at Layer} & \multicolumn{1}{l}{\multirow{2}{*}{\begin{tabular}[c]{@{}c@{}}\LineTAB\end{tabular}}} \\
                   & 1      & 4      & 8      & 12     & 16     & 20     & 24     & 28     & 32     & \multicolumn{1}{l}{} \\ \midrule
Iwslt2017 (BLEU)   & 37.94 & 38.99 & 39.47 & 38.68 & 38.10 & 35.33 & 33.24 & 28.90 & 22.40 & 41.37               \\
Squad (EM)         & 85.30 & 86.50 & 87.55 & 85.51 & 80.01 & 71.71 & 65.90 & 60.70 & 56.22 & 87.73               \\
GSM8K (EM)        & 13.04 & 19.33 & 19.56 & 14.93 & 11.67 & 6.14  & 4.92  & 3.56  & 2.80  & 27.07               \\
Hellaswag (EM)     & 77.70 & 88.36 & 87.92 & 85.75 & 85.46 & 77.44 & 27.96 & 49.52 & 47.79 & 91.76               \\
Dialogsum (RougeL) & 37.75 & 37.80 & 39.17 & 38.73 & 37.26 & 34.55 & 33.54 & 33.02 & 31.64 & 38.73               \\ \bottomrule
\end{tabular}
}
\caption{LoRA applied to a single layer in the transformer vs. \TAB. All results in this table are for a rank of $16$ for the 7B base model and use the same number of trainable parameters.}
\label{tab:layercheck}
\end{table*}
We can immediately see that \Lora is the best performing model for both the 2B and 7B base language models on most tasks.
This is hardly surprising as it is the most expensive method with respect to trainable parameters. 
The \TABuv configuration demonstrates the best overall performance among all \Tiedlora~configurations for both model sizes and is not far behind LoRA.
For our translation task with the LLaMA2 7B base model, \TABuv out performs \Lora while using $12.5\%$ of the number of parameters.
This consistent performance illustrates it effectiveness across diverse model scales and task types. 
The \TAB configuration, while marginally outperformed by \TABuv, is notable for its parameter efficiency. 
This method achieves comparable performance to \TABuv for typical ranks $r=8,16,32,64,128$, but with a reduced parameter count.

Both \TAB and \TABuv show robust performance at the same rank where traditional \Lora is optimized (i.e., performed best). 
This implies that for systems pre-tuned for \Lora, \TAB and \TABuv can be utilized with the same rank configuration as a ``drop-in'' replacement.

On average, we observe a $1.36$~\% decline in \TABuv performance compared to the \Lora model with the LLaMA 2 7B model. 
This decline, however, is marginally higher at $1.95$~\% with the 2B model. 
These findings suggest that the efficiency of \TABuv configuration may enhance with a larger or more capable base models (such as the LLaMA2 70B model). 
This hypothesis warrants a future exploration which we leave for future research.
\subsection{Task-Dependent Optimal Rank}
From~\Cref{tab:result}, we can see that the optimal rank for \Lora varies significantly across different tasks. 
Furthermore, perhaps surprisingly, a higher rank does not result in higher scores.
For example, for \Lora, a rank of 2 suffices for achieving best performance for the SQuAD task, while a higher rank of 64 is optimal for GSM8K.
In scenarios where traditional \Lora requires a higher rank, \tiedlora, especially \TAB and \TABuv, present an effective alternative by delivering comparable performance with substantially fewer parameters. For instance, for GSM8K, \TABuv needs only $4.3$~\% of the parameters that \Lora uses, while achieving a comparable performance (EM score of $31.77$ vs. $32.75$ for \TABuv and LoRA, respectively). 
\subsection{Layer Selection Vs. \Tiedlora}
The success of \Tiedlora, specifically \TAB as seen from  \Cref{tab:result}, begs the question -- Would adding LoRA to a single transformer layer lead to similar performance as \TAB?  
\TAB does not use layer-specific parameters (recall the \bluev~and \blueu~are frozen and set to $\mathbf{1}$) and has the same parameter count as applying \Lora to a single layer in the transformer model.
To examine this, we trained all tasks with LoRA applied to a single transformer layer's attention projection matrices. 
The obvious follow-up question is, which layer should LoRA be applied to?
We attempt single-layer LoRA on the lowest (closest to the input embeddings) layer, which we designate as ``Layer 1'' all the way to the the highest layer, ``Layer 32'' in 4 layer increments. We used the LLaMa2 7B model for this investigation.

\Cref{tab:layercheck} compares the performances of single-layer LoRA against \TAB (which uses the same number of trainable parameters as single-layer LoRA).
The \TAB configuration was considerably better than any of the layer selection LoRA settings.
Interestingly, we note that when applying LoRA to a single transformer layer, the lower layers (usually layer 4 or 8) resulted in higher performance than higher layers.
This suggests that there is potentially a single low-rank update that can be applied to all layers to boost performance, but it is hard to find a low-rank update for a single-layer that results in strong performance.
\subsection{Stability Across Ranks}
As indicated by~\Cref{subfig:all2b,subfig:all7b}, apart from \TABuv, all other \Tiedlora~methods experience a decline in performance when the rank increases. 
This trend highlights a general challenge faced by \Tiedlora~configurations, with \TABuv being an exception. 
Specifically, \TB and \TBu exhibit the most dramatic drop at higher ranks among all \Tiedlora\ configurations. 
We leave addressing these limitations for future research.

\Cref{subfig:best2b,subfig:best7b} show only the best \tiedlora~configurations, along with baselines \Lora and \Tuv.
While \TAB aligns closely with \TABuv at typical ranks of $4-16$, it also exhibits a small performance reduction at higher ranks. 
This pattern is repeated for \Tuv as well. 
In contrast, \TABuv maintains high performance across a broad range of ranks and is closest to \Lora.
    \newenvironment{customlegend}[1][]{%
        \begingroup
        \csname pgfplots@init@cleared@structures\endcsname
        \pgfplotsset{#1}%
    }{%
        \csname pgfplots@createlegend\endcsname
        \endgroup
    }%
    \def\addlegendimage{\csname pgfplots@addlegendimage\endcsname}

\pgfplotsset{ 
cycle list={%
{draw=black,mark=star,solid},
{draw=black, mark=square,solid}}}

 \definecolor{loracolor}{HTML}{66c2a5}
        \definecolor{tabcolor}{HTML}{fc8d62}
        \definecolor{tuvcolor}{HTML}{8da0cb}
        \definecolor{tacolor}{HTML}{e78ac3}
        \definecolor{tbcolor}{HTML}{e5c494}
	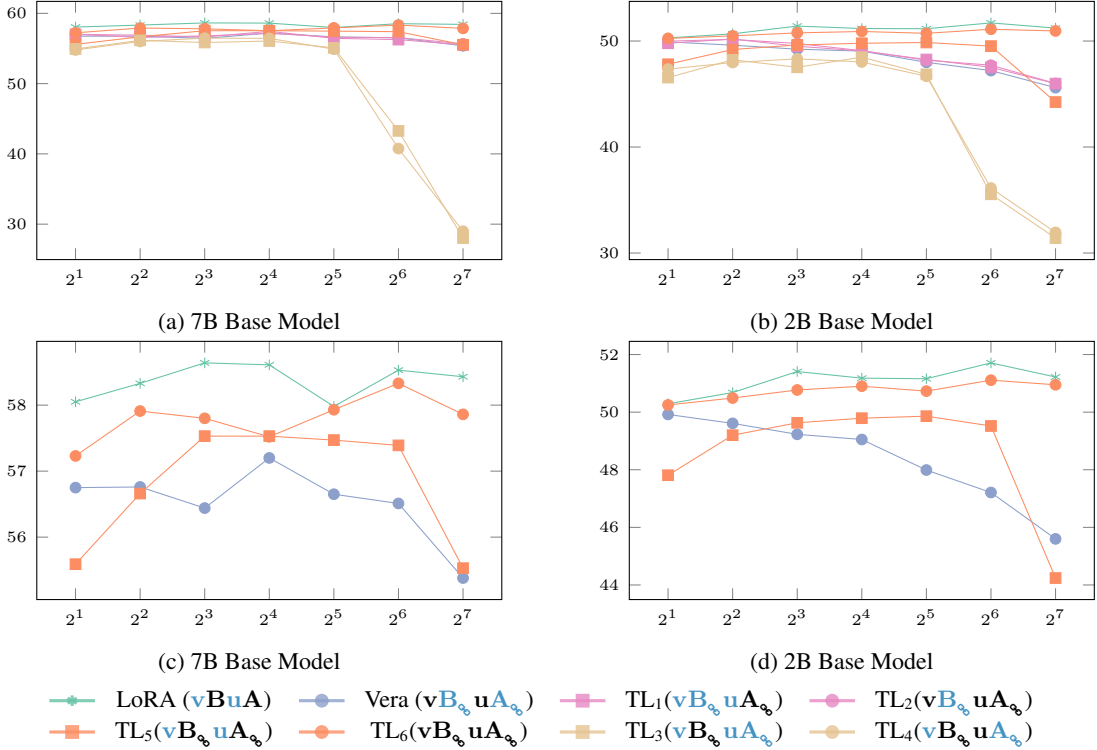
\begin{figure*}
	    \centering

    \subcaptionbox{7B Base Model\label{subfig:all7b}}
    [\columnwidth]{
	\begin{tikzpicture}[font=\tiny]
	    \begin{axis}[
                thin,
		height=0.65\columnwidth,
		width=\columnwidth,
                ylabel style={align=center},
		xmode=log,
		log basis y=2,
		log basis x=2,
		xtick={2,4,8,16,32,64,128}]
\addplot[mark=asterisk,loracolor]	coordinates	{	(2,58.05)	(4,58.33)	(8,58.64)	(16,58.61)	(32,57.98)	(64,58.53)	(128,58.43)	};
\addplot[mark=*,tuvcolor]	coordinates	{	(2,56.75)	(4,56.76)	(8,56.44)	(16,57.2)	(32,56.65)	(64,56.51)	(128,55.38)	};
\addplot[mark=square*,tacolor] 	coordinates	{	(2,57.01)	(4,56.9)	(8,56.63)	(16,57.44)	(32,56.45)	(64,56.26)	(128,55.52)	};
\addplot[mark=*,tacolor]	coordinates	{	(2,57.02)	(4,56.57)	(8,56.79)	(16,57.16)	(32,56.6)	(64,56.55)	(128,55.66)	};
\addplot[mark=square*,tabcolor]	coordinates	{	(2,55.59)	(4,56.66)	(8,57.53)	(16,57.53)	(32,57.47)	(64,57.39)	(128,55.53)	};
\addplot[mark=*,tabcolor]	coordinates	{	(2,57.23)	(4,57.91)	(8,57.8)	(16,57.52)	(32,57.93)	(64,58.33)	(128,57.86)	};
\addplot[mark=square*,tbcolor]	coordinates	{	(2,54.93)	(4,56.16)	(8,55.85)	(16,56.06)	(32,55.09)	(64,43.27)	(128,28.02)	};
\addplot[mark=*,tbcolor]	coordinates	{	(2,54.79)	(4,56.0)	(8,56.48)	(16,56.44)	(32,54.9)	(64,40.76)	(128,29.0)	};
	    \end{axis}
	\end{tikzpicture}
    }
    \subcaptionbox{2B Base Model\label{subfig:all2b}}
    [\columnwidth]{
	\begin{tikzpicture}[font=\tiny]
	    \begin{axis}[
                thin,
		height=0.65\columnwidth,
		width=\columnwidth,
                ylabel style={align=center},
		xmode=log,
		log basis y=2,
		log basis x=2,
		xtick={2,4,8,16,32,64,128}]
\addplot[mark=asterisk,loracolor]	coordinates	{	(2,50.29)	(4,50.68)	(8,51.41)	(16,51.18)	(32,51.16)	(64,51.71)	(128,51.22)	};
\addplot[mark=*,tuvcolor]	coordinates	{	(2,49.92)	(4,49.61)	(8,49.23)	(16,49.05)	(32,47.99)	(64,47.21)	(128,45.6)	};
\addplot[mark=square*,tacolor] 	coordinates	{	(2,49.79)	(4,50.19)	(8,49.51)	(16,49.06)	(32,48.26)	(64,47.52)	(128,45.98)	};
\addplot[mark=*,tacolor]	coordinates	{	(2,49.99)	(4,50.15)	(8,49.77)	(16,49.09)	(32,48.19)	(64,47.72)	(128,46.0)	};
\addplot[mark=square*,tabcolor]	coordinates	{	(2,47.81)	(4,49.2)	(8,49.63)	(16,49.79)	(32,49.86)	(64,49.52)	(128,44.24)	};
\addplot[mark=*,tabcolor]	coordinates	{	(2,50.25)	(4,50.49)	(8,50.77)	(16,50.9)	(32,50.73)	(64,51.11)	(128,50.95)	};
\addplot[mark=square*,tbcolor]	coordinates	{	(2,46.55)	(4,48.23)	(8,47.52)	(16,48.48)	(32,46.86)	(64,35.54)	(128,31.41)	};
\addplot[mark=*,tbcolor]	coordinates	{	(2,47.35)	(4,47.97)	(8,48.32)	(16,48.02)	(32,46.68)	(64,36.13)	(128,31.94)	};
	    \end{axis}
	\end{tikzpicture}
    }
    	    \subcaptionbox{7B Base Model\label{subfig:best7b}}
    [\columnwidth]{
	\begin{tikzpicture}[font=\tiny]
	    \begin{axis}[
                thin,
		height=0.65\columnwidth,
		width=\columnwidth,
                ylabel style={align=center},
		xmode=log,
		log basis x=2,
		xtick={2,4,8,16,32,64,128}]
\addplot[mark=asterisk,loracolor]	coordinates	{	(2,58.05)	(4,58.33)	(8,58.64)	(16,58.61)	(32,57.98)	(64,58.53)	(128,58.43)	};
\addplot[mark=*,tuvcolor]	coordinates	{	(2,56.75)	(4,56.76)	(8,56.44)	(16,57.2)	(32,56.65)	(64,56.51)	(128,55.38)	};
\addplot[mark=square*,tabcolor]	coordinates	{	(2,55.59)	(4,56.66)	(8,57.53)	(16,57.53)	(32,57.47)	(64,57.39)	(128,55.53)	};
\addplot[mark=*,tabcolor]	coordinates	{	(2,57.23)	(4,57.91)	(8,57.8)	(16,57.52)	(32,57.93)	(64,58.33)	(128,57.86)	};
	    \end{axis}
	\end{tikzpicture}
    }
    \subcaptionbox{2B Base Model\label{subfig:best2b}}
    [\columnwidth]{
	\begin{tikzpicture}[font=\tiny]
	    \begin{axis}[
                thin,
		height=0.65\columnwidth,
		width=\columnwidth,
                ylabel style={align=center},
		xmode=log,
		log basis x=2,
		xtick={2,4,8,16,32,64,128}]
\addplot[mark=asterisk,loracolor]	coordinates	{	(2,50.29)	(4,50.68)	(8,51.41)	(16,51.18)	(32,51.16)	(64,51.71)	(128,51.22)	};
\addplot[mark=*,tuvcolor]	coordinates	{	(2,49.92)	(4,49.61)	(8,49.23)	(16,49.05)	(32,47.99)	(64,47.21)	(128,45.6)	};
\addplot[mark=square*,tabcolor]	coordinates	{	(2,47.81)	(4,49.2)	(8,49.63)	(16,49.79)	(32,49.86)	(64,49.52)	(128,44.24)	};
\addplot[mark=*,tabcolor]	coordinates	{	(2,50.25)	(4,50.49)	(8,50.77)	(16,50.9)	(32,50.73)	(64,51.11)	(128,50.95)	};
	    \end{axis}
	\end{tikzpicture}
    }
\subcaptionbox*{}
    [\textwidth]{
    \begin{tikzpicture}[font=\tiny]
\begin{customlegend}[legend columns=4,legend style={align=left,draw=none,column sep=1ex},
        legend entries={\small{\lora},
                        \small{\Tuv},
                        \small{\TA},
                        \small{\TAuv},
                        \small{\TAB},
                        \small{\TABuv},
                        \small{\TB},
                        \small{\TBu},
                        }]
        \addlegendimage{opacity=0.8,very thick,mark=asterisk,loracolor}
        \addlegendimage{opacity=0.8,very thick,mark=*,tuvcolor}  
        \addlegendimage{opacity=0.8,very thick,mark=square*,tacolor}
        \addlegendimage{opacity=0.8,very thick,mark=*,tacolor}
        \addlegendimage{opacity=0.8,very thick,mark=square*,tabcolor}
        \addlegendimage{opacity=0.8,very thick,mark=*,tabcolor}
        \addlegendimage{opacity=0.8,very thick,mark=square*,tbcolor}
        \addlegendimage{opacity=0.8,very thick,mark=*,tbcolor}
        \end{customlegend}
\end{tikzpicture}
}
\vspace{-20pt}
	    \caption{Plots showing the performance of the \tiedlora configurations averaged over tasks across all ranks.\Cref{subfig:all2b,subfig:all7b} display all \tiedlora configurations, while \Cref{subfig:best2b,subfig:best7b} display the best \tiedlora configurations with LoRA and Vera as baselines.~\Cref{sec:detailedPlots} contains plots for each task and base model.}
	    \label{fig:plots}
	\end{figure*}

\section{Related Work}
\paragraph{Parameter-efficient fine-tuning (PEFT):} Recent work on PEFT of pretrained language models has shown competitive capabilities, often matching full fine-tuning performance for task-specific model customization while utilizing significantly fewer trainable parameters. 
Adapters~\citep{houlsby2019parameter, pfeiffer-etal-2021-adapterfusion} introduce task-specific parameters within the transformer layers that adapt to a particular task. 
Prompt tuning based methods such as P-Tuning and Prefix-Tuning~\citep{li-liang-2021-prefix,liu2023gpt} attempt to do the same but via task-specific vectors that can be appended to the inputs or at various layer representations.
BitFit and IA3~\citep{ben-zaken-etal-2022-bitfit,liu2022few} are PEFT methods that attempt to only alter bias vectors or scaling vectors in the base large language model.

\paragraph{Low-Rank adaptation (LoRA):} One of the most popular PEFT techniques is LoRA, introduced by ~\citet{hu2021lora}. LoRA employs low-rank matrix approximations of full weights' gradient-descent (GD) update to significantly reduce the number of trainable parameters. Importantly, LoRA can incorporate the low-rank updates into the frozen base weights after the fine-tuning process, avoiding any inference speed penalties or model architecture changes. In summary, LoRA paves the way for efficient fine-tuning for task-specific customization of large models with minimal computational overhead and no changes to the model's architecture.

\paragraph{Extensions to LoRA:} Since its arrival, there have been several efforts to improve the LoRA method. These methods mostly concentrated around reducing the trainable parameters and memory footprint while increasing the performance of the method on downstream tasks. AdaLoRA~\citep{zhang2023adaptive} introduces dynamic rank adjustment for the low-rank matrices during the fine-tuning process. The fundamental premise of this extension is to optimally distribute the parameter budget over model layers. \citet{chavan2023one} combined the adapter tuning with LoRA to derive a generalized framework that utilized both methods for increased flexibility and capability across a wide variety of tasks and datasets. \citet{kopiczko2023vera} proposes the VeRA method the freezes randomly initialized projection matrices and introduces trainable scaling vectors that vary across layers. This method shows similar performance to the \lora method while dramatically reducing the number of trainable parameters. We view VeRA as one specific configuration which lies on end of the \Tiedlora~spectrum. 

Tangential to the efforts that aim to reduce trainable parameters, QLoRA~\citep{dettmers2023qlora}, significantly reduces the memory usage of LoRA using a 4-bit or 8-bit quantized base language model during training. The method provides algorithms and custom kernels to backpropagate gradients through the frozen, quantized base model to update low-rank matrices during training, resulting in considerable reduction in memory usage. Combining quantization and reduction in the number of trainable parameters is a direction of future work.
\paragraph{Weight tying:}  Weight tying~\citep{press-wolf-2017-using,inan2016tying} is a common approach that reduces the number of parameters  by using the same set of weights in different parts of the network. Typically the input word embedding layer and the output word embedding layer (sometimes referred to as the language model head) are tied.
In this study, we apply weight tying to the low-rank weight matrices used in LoRA, and share them across the layers of the base language model. 
This simple procedure leads to efficient training methods where the number of trainable parameters are either unaffected by, or only increases marginally with the number of hidden layers. As models get deeper this approach naturally provides greater parameter reduction over original LoRA method.
\paragraph{Vision Transformers:} Ideas similar to Tied-Lora are also being explored in the vision based tasks. \citet{dong2023efficient}, for example, uses weight tying and bottleneck adapters.
\section{Conclusion \& Future Work}
In this paper, we introduced the \TiedLora~paradigm, a novel approach to enhance the parameter efficiency of Lora by employing a simple technique of weight-tying and selective training of low-rank matrices.

Our empirical analysis demonstrates that the \TABuv~configuration achieves performance comparable to Lora across various tasks, while utilizing only a fraction of the parameters employed by Lora across a spectrum of low-rank dimensions. This efficiency becomes more pronounced at higher ranks, leading to a more aggressive reduction in the number of trainable parameters compared to Lora. Remarkably, in the translation task, \TABuv~surpassed Lora's performance while using only $12.5\%$ of the number of parameters.
Our study highlights that the benefits of this configuration are particularly evident in tasks that leverage the inherent strengths of the base language model, such as commonsense NLI, extractive QA, and summarization. Tasks involving mathematical reasoning and arithmetic calculations, however, favor the sheer learning capacity of LoRA with more parameters.

As language models continue to advance, the \TiedLora~configurations, with their optimized efficiency, emerge as a promising candidate to replace traditional Lora in a broader range of applications. This progression underscores the relevance of \TiedLora~as a scalable solution in the dynamic landscape of large language model customization.

For future research, we plan to delve into the application of \TiedLora~methods on larger base models. This exploration aims to assess their scalability and effectiveness within the broader context of large language models. Additionally, we intend to investigate weight tying in other parameter-efficient fine-tuning methods such as Adapters and Prefix Tuning, both of which introduce layer-specific parameters.

\section*{Limitations}
PEFT methods are inherently sensitive to the base large language which they are applied to as well as the specific customization task. While we attempt to test our methods on multiple base models computation cost restricts us to only $2$ models (so far). While extending our analysis to other models is possible, extending to more tasks is more challenging as the variety of tasks is large. 
Furthermore, predicting the behavior of a PEFT method on a new task based on its performance on some existing task is very challenging. Even in our analysis we did not expect translation task to be an outlier (our \Tiedlora~method out performed LoRA) because on all the other tasks LoRA was slightly better. Thus, we caution against very strong claims of task generalization and highlight that while we show results on diverse tasks there are still a wide range of tasks we have not explored.
\bibliography{anthology,custom}
\bibliographystyle{acl_natbib}
\clearpage
\appendix
\section{Breakdown of Performances with Ranks}\label{sec:detailedPlots}

\pgfplotsset{ 
cycle list={%
{draw=black,mark=star,solid},
{draw=black, mark=square,solid}}}

        \definecolor{loracolor}{HTML}{66c2a5}
        \definecolor{tabcolor}{HTML}{fc8d62}
        \definecolor{tuvcolor}{HTML}{8da0cb}
        \definecolor{tacolor}{HTML}{e78ac3}
        \definecolor{tbcolor}{HTML}{e5c494}
        \begin{figure*}
            \centering
            \subcaptionbox{iwslt2017,2B}
    [\columnwidth]{
        \vspace{20pt}
        \begin{tikzpicture}[font=\tiny]
            \begin{axis}[
                thick,
                height=0.5\columnwidth,
                width=\columnwidth,
                ylabel style={align=center},
                ylabel=iwslt2017\\BLEU,
                ymin=37.119,
                xmode=log,
                log basis x=2,
                xtick={2,4,8,16,32,64,128}]
                \addplot[mark=asterisk,loracolor] coordinates {(2,39.325) (4,39.674) (8,39.849) (16,39.849) (32,39.952) (64,39.968) (128,40.193)};
\addplot[mark=*,tuvcolor] coordinates {(2,39.152) (4,38.872) (8,38.874) (16,38.794) (32,38.569) (64,38.612) (128,38.204)};
\addplot[mark=square*,tacolor] coordinates {(2,39.095) (4,38.843) (8,38.896) (16,38.825) (32,38.885) (64,38.702) (128,38.119)};
\addplot[mark=*,tacolor] coordinates {(2,39.108) (4,38.866) (8,38.94) (16,38.842) (32,38.618) (64,39.049) (128,38.175)};
\addplot[mark=square*,tabcolor] coordinates {(2,38.184) (4,38.491) (8,38.288) (16,39.046) (32,38.843) (64,39.203) (128,38.584)};
\addplot[mark=*,tabcolor] coordinates {(2,39.206) (4,39.411) (8,39.56) (16,39.609) (32,39.178) (64,39.596) (128,39.741)};
\addplot[mark=square*,tbcolor] coordinates {(2,37.524) (4,38.249) (8,37.964) (16,38.221) (32,37.115) (64,34.985) (128,36.464)};
\addplot[mark=*,tbcolor] coordinates {(2,37.687) (4,37.57) (8,38.303) (16,38.108) (32,37.341) (64,37.567) (128,32.978)};
            \end{axis}
        \end{tikzpicture}
    }
    \subcaptionbox{iwslt2017,7B}
    [\columnwidth]{
        \vspace{20pt}
        \begin{tikzpicture}[font=\tiny]
            \begin{axis}[
                thick,
                height=0.5\columnwidth,
                width=\columnwidth,
                ylabel style={align=center},
                ylabel=iwslt2017\\BLEU,
                ymin=38.821,
                xmode=log,
                log basis x=2,
                xtick={2,4,8,16,32,64,128}]
                \addplot[mark=asterisk,loracolor] coordinates {(2,40.704) (4,41.07) (8,41.304) (16,41.287) (32,40.683) (64,40.539) (128,40.778)};
\addplot[mark=*,tuvcolor] coordinates {(2,40.018) (4,40.101) (8,40.221) (16,40.415) (32,40.223) (64,40.354) (128,39.821)};
\addplot[mark=square*,tacolor] coordinates {(2,40.043) (4,40.234) (8,40.336) (16,40.472) (32,40.276) (64,40.525) (128,40.471)};
\addplot[mark=*,tacolor] coordinates {(2,40.032) (4,40.236) (8,40.352) (16,40.421) (32,40.291) (64,40.49) (128,40.497)};
\addplot[mark=square*,tabcolor] coordinates {(2,39.847) (4,40.364) (8,40.007) (16,41.37) (32,40.679) (64,40.641) (128,40.512)};
\addplot[mark=*,tabcolor] coordinates {(2,40.188) (4,41.04) (8,41.334) (16,40.015) (32,41.374) (64,40.381) (128,40.497)};
\addplot[mark=square*,tbcolor] coordinates {(2,39.364) (4,39.525) (8,39.927) (16,40.479) (32,40.24) (64,38.357) (128,6.903)};
\addplot[mark=*,tbcolor] coordinates {(2,39.228) (4,39.678) (8,40.335) (16,40.653) (32,40.454) (64,38.879) (128,5.96)};
            \end{axis}
        \end{tikzpicture}
    }
    \subcaptionbox{squad,2B}
    [\columnwidth]{
        \vspace{20pt}
        \begin{tikzpicture}[font=\tiny]
            \begin{axis}[
                thick,
                height=0.5\columnwidth,
                width=\columnwidth,
                ylabel style={align=center},
                ylabel=squad\\Acc.,
                ymin=76.351,
                xmode=log,
                log basis x=2,
                xtick={2,4,8,16,32,64,128}]
                \addplot[mark=asterisk,loracolor] coordinates {(2,82.63) (4,82.914) (8,83.084) (16,83.283) (32,83.576) (64,83.349) (128,83.463)};
\addplot[mark=*,tuvcolor] coordinates {(2,81.769) (4,81.731) (8,80.473) (16,79.896) (32,79.432) (64,78.051) (128,78.079)};
\addplot[mark=square*,tacolor] coordinates {(2,82.204) (4,81.921) (8,81.145) (16,80.18) (32,79.262) (64,77.758) (128,77.871)};
\addplot[mark=*,tacolor] coordinates {(2,81.996) (4,82.412) (8,81.4) (16,79.981) (32,79.499) (64,77.985) (128,77.9)};
\addplot[mark=square*,tabcolor] coordinates {(2,80.407) (4,81.353) (8,82.081) (16,82.744) (32,81.429) (64,80.643) (128,77.351)};
\addplot[mark=*,tabcolor] coordinates {(2,82.573) (4,83.254) (8,83.046) (16,83.236) (32,83.018) (64,83.557) (128,83.453)};
\addplot[mark=square*,tbcolor] coordinates {(2,79.527) (4,80.956) (8,77.436) (16,80.596) (32,76.963) (64,75.695) (128,55.667)};
\addplot[mark=*,tbcolor] coordinates {(2,80.454) (4,80.908) (8,81.031) (16,78.903) (32,77.465) (64,74.588) (128,65.781)};
            \end{axis}
        \end{tikzpicture}
    }
    \subcaptionbox{squad,7B}
    [\columnwidth]{
        \vspace{20pt}
        \begin{tikzpicture}[font=\tiny]
            \begin{axis}[
                thick,
                height=0.5\columnwidth,
                width=\columnwidth,
                ylabel style={align=center},
                ylabel=squad\\Acc.,
                ymin=83.986,
                xmode=log,
                log basis x=2,
                xtick={2,4,8,16,32,64,128}]
                \addplot[mark=asterisk,loracolor] coordinates {(2,88.524) (4,88.307) (8,88.231) (16,88.344) (32,88.174) (64,87.985) (128,88.448)};
\addplot[mark=*,tuvcolor] coordinates {(2,87.692) (4,87.398) (8,87.569) (16,86.991) (32,87.304) (64,86.585) (128,85.79)};
\addplot[mark=square*,tacolor] coordinates {(2,87.72) (4,87.531) (8,87.512) (16,87.02) (32,86.963) (64,86.537) (128,85.033)};
\addplot[mark=*,tacolor] coordinates {(2,87.673) (4,87.37) (8,87.483) (16,86.878) (32,86.812) (64,86.613) (128,86.102)};
\addplot[mark=square*,tabcolor] coordinates {(2,87.114) (4,88.221) (8,87.465) (16,87.739) (32,87.502) (64,87.067) (128,84.986)};
\addplot[mark=*,tabcolor] coordinates {(2,87.975) (4,88.486) (8,87.72) (16,88.042) (32,88.127) (64,88.202) (128,88.467)};
\addplot[mark=square*,tbcolor] coordinates {(2,87.342) (4,87.465) (8,87.616) (16,86.944) (32,86.121) (64,82.1) (128,58.061)};
\addplot[mark=*,tbcolor] coordinates {(2,87.058) (4,87.72) (8,87.408) (16,87.247) (32,86.093) (64,82.838) (128,77.871)};
            \end{axis}
        \end{tikzpicture}
    }
    \subcaptionbox{gsm8k,2B}
    [\columnwidth]{
        \vspace{20pt}
        \begin{tikzpicture}[font=\tiny]
            \begin{axis}[
                thick,
                height=0.5\columnwidth,
                width=\columnwidth,
                ylabel style={align=center},
                ylabel=gsm8k\\Acc.,
                ymin=2.791,
                xmode=log,
                log basis x=2,
                xtick={2,4,8,16,32,64,128}]
                \addplot[mark=asterisk,loracolor] coordinates {(2,9.477) (4,8.567) (8,11.979) (16,11.145) (32,11.069) (64,12.282) (128,10.993)};
\addplot[mark=*,tuvcolor] coordinates {(2,8.264) (4,7.051) (8,6.899) (16,7.354) (32,6.596) (64,6.975) (128,4.397)};
\addplot[mark=square*,tacolor] coordinates {(2,8.112) (4,9.553) (8,7.961) (16,7.657) (32,6.823) (64,8.264) (128,4.246)};
\addplot[mark=*,tacolor] coordinates {(2,8.567) (4,9.401) (8,8.036) (16,7.43) (32,6.748) (64,8.112) (128,4.321)};
\addplot[mark=square*,tabcolor] coordinates {(2,3.791) (4,5.914) (8,7.051) (16,7.354) (32,7.885) (64,8.34) (128,9.401)};
\addplot[mark=*,tabcolor] coordinates {(2,8.643) (4,8.567) (8,10.159) (16,10.311) (32,8.491) (64,9.78) (128,9.856)};
\addplot[mark=square*,tbcolor] coordinates {(2,3.033) (4,4.776) (8,6.975) (16,6.065) (32,6.596) (64,5.686) (128,5.08)};
\addplot[mark=*,tbcolor] coordinates {(2,3.867) (4,4.018) (8,6.52) (16,6.52) (32,7.202) (64,6.823) (128,3.033)};
            \end{axis}
        \end{tikzpicture}
    }
    \subcaptionbox{gsm8k,7B}
    [\columnwidth]{
        \vspace{20pt}
        \begin{tikzpicture}[font=\tiny]
            \begin{axis}[
                thick,
                height=0.5\columnwidth,
                width=\columnwidth,
                ylabel style={align=center},
                ylabel=gsm8k\\Acc.,
                ymin=20.152,
                xmode=log,
                log basis x=2,
                xtick={2,4,8,16,32,64,128}]
                \addplot[mark=asterisk,loracolor] coordinates {(2,29.492) (4,30.023) (8,31.16) (16,31.16) (32,30.023) (64,32.752) (128,31.615)};
\addplot[mark=*,tuvcolor] coordinates {(2,27.218) (4,27.672) (8,25.777) (16,29.113) (32,26.763) (64,27.218) (128,25.171)};
\addplot[mark=square*,tacolor] coordinates {(2,28.127) (4,27.369) (8,27.142) (16,29.947) (32,25.701) (64,27.066) (128,25.625)};
\addplot[mark=*,tacolor] coordinates {(2,28.734) (4,26.839) (8,27.369) (16,28.734) (32,26.914) (64,27.066) (128,24.943)};
\addplot[mark=square*,tabcolor] coordinates {(2,21.152) (4,23.503) (8,28.127) (16,27.066) (32,28.506) (64,30.326) (128,29.947)};
\addplot[mark=*,tabcolor] coordinates {(2,26.611) (4,28.734) (8,30.023) (16,28.658) (32,28.81) (64,31.766) (128,29.037)};
\addplot[mark=square*,tbcolor] coordinates {(2,19.864) (4,24.337) (8,21.228) (16,24.033) (32,23.199) (64,17.741) (128,15.466)};
\addplot[mark=*,tbcolor] coordinates {(2,18.878) (4,22.82) (8,25.701) (16,24.64) (32,23.199) (64,21.001) (128,15.466)};
            \end{axis}
        \end{tikzpicture}
    }
    \subcaptionbox{hellaswag,2B}
    [\columnwidth]{
        \vspace{20pt}
        \begin{tikzpicture}[font=\tiny]
            \begin{axis}[
                thick,
                height=0.5\columnwidth,
                width=\columnwidth,
                ylabel style={align=center},
                ylabel=hellaswag\\Acc.,
                ymin=58.221,
                xmode=log,
                log basis x=2,
                xtick={2,4,8,16,32,64,128}]
                \addplot[mark=asterisk,loracolor] coordinates {(2,82.733) (4,83.679) (8,84.515) (16,84.266) (32,84.236) (64,85.64) (128,85.132)};
\addplot[mark=*,tuvcolor] coordinates {(2,83.41) (4,83.39) (8,82.623) (16,82.643) (32,78.759) (64,75.941) (128,70.464)};
\addplot[mark=square*,tacolor] coordinates {(2,82.703) (4,83.539) (8,82.314) (16,82.195) (32,79.735) (64,76.319) (128,72.834)};
\addplot[mark=*,tacolor] coordinates {(2,83.639) (4,83.081) (8,83.171) (16,82.663) (32,79.496) (64,77.017) (128,72.675)};
\addplot[mark=square*,tabcolor] coordinates {(2,80.343) (4,83.091) (8,83.191) (16,83.001) (32,83.708) (64,82.245) (128,59.221)};
\addplot[mark=*,tabcolor] coordinates {(2,83.37) (4,83.579) (8,83.609) (16,84.356) (32,85.132) (64,85.023) (128,84.525)};
\addplot[mark=square*,tbcolor] coordinates {(2,76.29) (4,80.661) (8,80.074) (16,80.353) (32,77.435) (64,25.045) (128,25.045)};
\addplot[mark=*,tbcolor] coordinates {(2,79.805) (4,80.512) (8,78.859) (16,79.994) (32,74.497) (64,25.045) (128,25.045)};
            \end{axis}
        \end{tikzpicture}
    }
    \subcaptionbox{hellaswag,7B}
    [\columnwidth]{
        \vspace{20pt}
        \begin{tikzpicture}[font=\tiny]
            \begin{axis}[
                thick,
                height=0.5\columnwidth,
                width=\columnwidth,
                ylabel style={align=center},
                ylabel=hellaswag\\Acc.,
                ymin=82.489,
                xmode=log,
                log basis x=2,
                xtick={2,4,8,16,32,64,128}]
                \addplot[mark=asterisk,loracolor] coordinates {(2,91.316) (4,91.675) (8,91.725) (16,91.974) (32,91.127) (64,91.068) (128,91.775)};
\addplot[mark=*,tuvcolor] coordinates {(2,90.47) (4,89.912) (8,89.843) (16,89.912) (32,89.544) (64,88.349) (128,86.805)};
\addplot[mark=square*,tacolor] coordinates {(2,90.301) (4,90.52) (8,89.414) (16,90.032) (32,90.042) (64,87.941) (128,87.074)};
\addplot[mark=*,tacolor] coordinates {(2,90.321) (4,89.892) (8,90.072) (16,90.112) (32,89.693) (64,88.767) (128,87.423)};
\addplot[mark=square*,tabcolor] coordinates {(2,90.898) (4,91.426) (8,91.436) (16,91.755) (32,90.848) (64,89.504) (128,83.489)};
\addplot[mark=*,tabcolor] coordinates {(2,91.735) (4,91.824) (8,90.669) (16,91.147) (32,91.705) (64,91.904) (128,91.814)};
\addplot[mark=square*,tbcolor] coordinates {(2,89.484) (4,90.171) (8,90.271) (16,89.375) (32,86.766) (64,39.016) (128,25.045)};
\addplot[mark=*,tbcolor] coordinates {(2,89.823) (4,90.181) (8,89.474) (16,89.464) (32,85.411) (64,25.045) (128,24.846)};
            \end{axis}
        \end{tikzpicture}
    }
    \subcaptionbox{dialogsum,2B}
    [\columnwidth]{
        \vspace{20pt}
        \begin{tikzpicture}[font=\tiny]
            \begin{axis}[
                thick,
                height=0.5\columnwidth,
                width=\columnwidth,
                ylabel style={align=center},
                ylabel=dialogsum\\RougeL,
                ymin=35.328,
                xmode=log,
                log basis x=2,
                xtick={2,4,8,16,32,64,128}]
                \addplot[mark=asterisk,loracolor] coordinates {(2,37.283) (4,38.586) (8,37.632) (16,37.369) (32,36.973) (64,37.316) (128,36.331)};
\addplot[mark=*,tuvcolor] coordinates {(2,37.006) (4,37.015) (8,37.277) (16,36.579) (32,36.614) (64,36.47) (128,36.872)};
\addplot[mark=square*,tacolor] coordinates {(2,36.812) (4,37.106) (8,37.221) (16,36.429) (32,36.599) (64,36.553) (128,36.809)};
\addplot[mark=*,tacolor] coordinates {(2,36.646) (4,36.997) (8,37.29) (16,36.509) (32,36.592) (64,36.44) (128,36.936)};
\addplot[mark=square*,tabcolor] coordinates {(2,36.328) (4,37.173) (8,37.55) (16,36.802) (32,37.412) (64,37.164) (128,36.635)};
\addplot[mark=*,tabcolor] coordinates {(2,37.473) (4,37.63) (8,37.48) (16,37.003) (32,37.809) (64,37.605) (128,37.165)};
\addplot[mark=square*,tbcolor] coordinates {(2,36.358) (4,36.5) (8,35.161) (16,37.176) (32,36.166) (64,36.264) (128,34.793)};
\addplot[mark=*,tbcolor] coordinates {(2,34.922) (4,36.818) (8,36.863) (16,36.552) (32,36.885) (64,36.617) (128,32.852)};
            \end{axis}
        \end{tikzpicture}
    }
    \subcaptionbox{dialogsum,7B}
    [\columnwidth]{
        \vspace{20pt}
        \begin{tikzpicture}[font=\tiny]
            \begin{axis}[
                thick,
                height=0.5\columnwidth,
                width=\columnwidth,
                ylabel style={align=center},
                ylabel=dialogsum\\RougeL,
                ymin=37.348,
                xmode=log,
                log basis x=2,
                xtick={2,4,8,16,32,64,128}]
                \addplot[mark=asterisk,loracolor] coordinates {(2,40.232) (4,40.596) (8,40.757) (16,40.287) (32,39.896) (64,40.295) (128,39.523)};
\addplot[mark=*,tuvcolor] coordinates {(2,38.361) (4,38.73) (8,38.775) (16,39.592) (32,39.417) (64,40.066) (128,39.316)};
\addplot[mark=square*,tacolor] coordinates {(2,38.856) (4,38.84) (8,38.727) (16,39.744) (32,39.259) (64,39.24) (128,39.382)};
\addplot[mark=*,tacolor] coordinates {(2,38.348) (4,38.497) (8,38.687) (16,39.657) (32,39.269) (64,39.815) (128,39.352)};
\addplot[mark=square*,tabcolor] coordinates {(2,38.934) (4,39.797) (8,40.625) (16,39.731) (32,39.832) (64,39.4) (128,38.728)};
\addplot[mark=*,tabcolor] coordinates {(2,39.624) (4,39.483) (8,39.243) (16,39.713) (32,39.657) (64,39.399) (128,39.497)};
\addplot[mark=square*,tbcolor] coordinates {(2,38.605) (4,39.287) (8,40.203) (16,39.479) (32,39.148) (64,39.115) (128,34.602)};
\addplot[mark=*,tbcolor] coordinates {(2,38.961) (4,39.624) (8,39.461) (16,40.171) (32,39.341) (64,36.056) (128,20.88)};
            \end{axis}
        \end{tikzpicture}
    }
\subcaptionbox*{}
    [\textwidth]{
     \vspace{10pt}
    \begin{tikzpicture}
\begin{customlegend}[legend columns=4,legend style={align=left,draw=none,column sep=2ex},
        legend entries={\lora ,
                        \Tuv ,
                        \TA,
                        \TAuv ,
                        \TAB ,
                        \TABuv ,
                        \TB ,
                        \TBu,
                        }]
        \addlegendimage{opacity=0.8,very thick,mark=asterisk,loracolor}
        \addlegendimage{opacity=0.8,very thick,mark=*,tuvcolor}  
        \addlegendimage{opacity=0.8,very thick,mark=square*,tacolor}
        \addlegendimage{opacity=0.8,very thick,mark=*,tacolor}
        \addlegendimage{opacity=0.8,very thick,mark=square*,tabcolor}
        \addlegendimage{opacity=0.8,very thick,mark=*,tabcolor}
        \addlegendimage{opacity=0.8,very thick,mark=square*,tbcolor}
        \addlegendimage{opacity=0.8,very thick,mark=*,tbcolor}
        \end{customlegend}
\end{tikzpicture}
}
    \caption{Plots showing the performance of the \tiedlora~configurations along with the baseline \Lora for $5$ diverse tasks at $4$ different values for low-rank dimension setting. Note that we let the plot for \TB and \TBu go out of bounds to show details for the other curves.}
    \label{fig:detailplots}
\end{figure*}

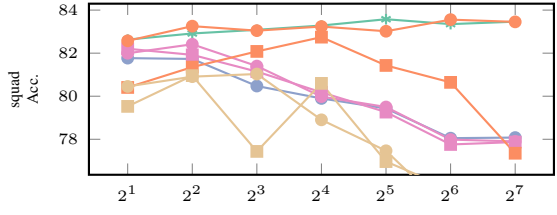
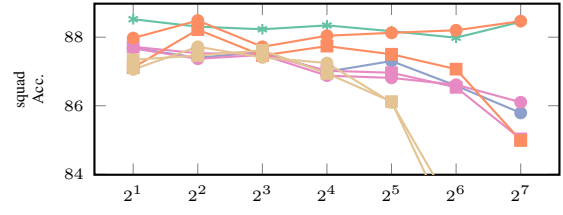
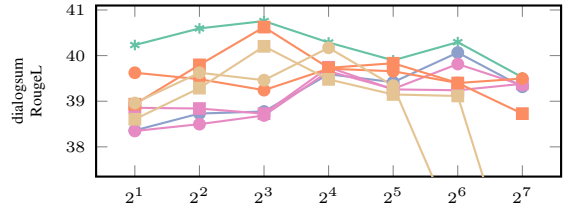
\end{document}